\documentclass{article}
\usepackage[affil-it]{authblk}
\usepackage{geometry}
\usepackage{url}
\usepackage{amsmath,amssymb,amsfonts,graphicx,amsthm,nicefrac,mathtools,bbm,bm}
\usepackage{hyperref}
\usepackage{verbatim}
\usepackage[numbers]{natbib}
\usepackage{algpseudocode}
\usepackage{booktabs,multirow}
\usepackage{fixltx2e}

\title{Graphical Exponential Screening}

\author{Zhe Liu}
\affil{University of Chicago}
\date{}

\newtheorem{theorem}{Theorem}[section]
\newtheorem{assumption}{Assumption}[section]
\newtheorem{definition}{Definition}[section]
\newtheorem{algorithm}{Algorithm}[section]
\newtheorem{proposition}{Proposition}[section]

\renewcommand\hat\widehat

\algdef{SE}[DOWHILE]{Do}{doWhile}{\algorithmicdo}[1]{\algorithmicwhile\ #1}%
\DeclareMathOperator{\tr}{\text{tr}}

\DeclareMathOperator{\KL}{\text{KL}}

\begin{document}
\maketitle

\begin{abstract}
In high dimensions we propose and analyze an aggregation estimator of the precision matrix for Gaussian graphical models. This estimator, called \emph{graphical Exponential Screening} (gES), linearly combines a suitable set of individual estimators with different underlying graphs, and balances the estimation error and sparsity. We study the risk of this aggregation estimator and show that it is comparable to that of the best estimator based on a single graph, chosen by an oracle. Numerical performance of our method is investigated using both simulated and real datasets, in comparison with some state-of-art estimation procedures.
\end{abstract}

\section{Introduction}
\label{intro}
Graphical models \citep{Lauritzen96, Whittaker09} have become as a useful way of exploring and modeling the distribution. For instance, graphical models could be used to represent complex interactions among gene products resulted from biological processes. Such problems require us to infer an undirected graph from i.i.d. observations.

Let $X=(X_1,\ldots,X_p)^T$ be a random vector with some continuous distribution. An undirected graph $G$ has $p$ vertices, collected in a set $V$, one for each variable. We represent the edges as a set $E$ of unordered pairs: $(i,j)\in E$ if and only if there is an edge between $X_i$ and $X_j$. An edge between $X_i$ and $X_j$ is absent if $X_i$ and $X_j$ are independent, given the other variables.

The default model for graphical modeling of continuous data is the multivariate Gaussian. Let data $\mathcal{D}_n:=\{x_1,\ldots,x_n\}$ be the realizations of $n$ independent samples from a multivariate Gaussian distribution $\mathcal{N}_p(0,\Sigma)$, where $\Sigma$ is the covariance matrix. Then the log-likelihood of $\mathcal{D}_n$ is (up to a constant) given by
\begin{align}
l_n(\Theta):=\frac{n}{2}\log\det(\Theta)-\frac{n}{2}\text{tr}(\widehat{\Sigma}_n\Theta),
\end{align}
where $\Theta=\Sigma^{-1}$ is the precision matrix, i.e. inverse covariance matrix, and $\widehat{\Sigma}_n=\frac{1}{n}\sum_{i=1}^n x_i x_i^T$ is the empirical covariance matrix. 

For Gaussian graphical models, it is well known that the edge between the $i$th and $j$th nodes is absent in the graph, meaning that the associated variables  are conditionally independent given the other variables, if and only if $\theta_{ij}=0$, where $\theta_{ij}$ is the $(i,j)$th element of $\Theta$. Therefore, the estimation and model selection problems in Gaussian graphical models are equivalent to estimation of the precision matrix and identification of its zero-pattern.

While it is one of the classical problems in multivariate statistics, with a renewed focus on high-dimensional data in recent years, a number of sparse estimators have been proposed to deal with the problem of precision matrix estimation. Among them, \citet{Yuan07} and \citet{Friedman08} impose an $\ell_1$ penalty on the entries of the precision matrix when maximizing the Gaussian log-likelihood, known as the \emph{graphical lasso}, encouraging some of the entries of the estimated precision matrix to be exactly zero. \citet{Meinshausen06} consider the neighborhood selection method via the lasso. \citet{Cai11} propose a constrained $\ell_1$ minimization approach for sparse inverse covariance matrix estimation. \citet{Yuan10} takes advantage of the connection between multivariate linear regression and entries of the inverse covariance matrix, developing an estimating procedure that can effectively exploit sparsity. Theoretical properties, including consistency in parameter estimation and sparsity structure recovery, are discussed in these and other papers \citep{Raskutti08, Rothman08}.

Given a collected family of estimators, linear or convex aggregation methods are another class of technique to address model selection problems and provide flexible ways to combine various models into a single estimator \citep{Rigollet11}. The idea of aggregating estimators was originally described in \citep{Nemirovski00}. The suggestion put forward by \citep{Nemirovski00} is to achieve two independent subsamples from the original sample by randomization: individual estimators are constructed from the first subsample while the second is used to perform aggregation on those individual estimators. This idea of two-step procedures carries over to models with i.i.d. observations where one can do sample splitting. Along with this method, one might aggregate estimators using the same observations for both estimation and aggregation. However, this would generally result in overfitting.

A primary motivation for aggregating estimators is that it can improve the estimation risk, as ``betting'' on multiple models can provide a type of insurance against a single model being poor \citep{Leung06}. Most of the recent work on estimator aggregation deals with regression learning problems. For example, \emph{Exponential Screening} (ES) for linear models provides a form of frequentist averaging over a large model class, which enjoys strong theoretical properties \citep{Rigollet11}. Also, an aggregation classifier is proposed in \citep{Lecue07} and an optimal rate of convex aggregation for the hinge risk is also obtained.

In this paper, we propose a new estimating procedure by considering a linear combination of a suitable set of individual estimators with different sparsity patterns. A sparsity pattern is defined as a binary vector with each element indicating whether the corresponding edge of the graph is absent or not. These individual estimators and the corresponding aggregating weights are determined to ensure a competitive rate of convergence for the risk of the aggregation estimator.

Our aggregation method is based on a sample-splitting procedure: the first subsample is set to construct individual estimators and the second subsample is then used to determine the weights and aggregate these estimators. To carry out the analysis of the aggregation step, it is enough to work conditionally on the first subsample so that the problem reduces to aggregation of deterministic estimators \citep{Rigollet12}. Namely, let $R(\cdot)$ denote a risk function, then given the deterministic estimator $\Theta_m$ with sparsity pattern $m$, one can construct an aggregation estimator $\widehat{\Theta}$ such that the excess risk
\begin{align}
r:=R(\widehat{\Theta})- \min_{m\in\mathcal{M}}R(\Theta_m),
\end{align}
is as small as possible, where $\mathcal{M}$ is a candidate set of sparsity patterns. Ideally, we wish to find an aggregation estimator whose risk is as close as possible (in a probabilistic sense) to the minimum risk of individual estimators. The risk function considered in the paper is the Kullback-Leibler divergence.

The rest of the paper is organized as follows. In Section~\ref{method}, we describe the graphical aggregation in detail. Theoretical properties are provided in Section~\ref{theory}. Numerical experiments are presented in Sections~\ref{num}. Finally, we conclude in Section~\ref{conclude}. Detailed proofs are collected in Appendix section.

\section{Graphical Exponential Screening Estimator}
\label{method}

\subsection{Notations and preliminaries}
Let $|\cdot|_1$ denote the $\ell_1$ vector norm, $\|\cdot\|$ denote the spectral matrix norm, and $\|\cdot\|_F$ denote the Frobenius matrix norm. For any symmetric positive-definite matrix $A$ we use the notation $A\succ 0$. For any two real numbers $a$ and $b$, we use the notation $a \vee b:=\max(a,b)$. Let $\mathbf{1}(\cdot)$ denote the indicator function.

We call a \emph{sparsity pattern} any binary vector $m\in\mathcal{M}$, where $\mathcal{M}$ is a candidate set of sparsity patterns and $\mathcal{M}\subset\{0, 1\}^{p(p-1)/2}$. The $k$th coordinate of $m$ can be interpreted as an indicator of presence ($m_k = 1$) or absence ($m_k = 0$) of the edge $(i_k,j_k)\in E$ such that $(i_k,j_k)$ is the $k$th element of the ordered list $\mathcal{S}_p$, where $\mathcal{S}_p=\{(i,j):1\leq i<j \leq p\}$.

We partition the sample $\mathcal{D}_n$ into two independent subsamples, $\mathcal{D}_{n_1}^{(1)}$ and $\mathcal{D}_{n_2}^{(2)}$, of size $n_1$ and $n_2$, respectively, where $n_1+n_2=n$.

\subsection{Graphical aggregation method}
In the aggregation procedure, $\mathcal{D}_{n_1}^{(1)}$ is utilized to construct individual estimators and $\mathcal{D}_{n_2}^{(2)}$ is then used to aggregate these estimators. Given each sparsity pattern $m\in\mathcal{M}$, we first define the individual estimator of the precision matrix.

\begin{definition} For each $m\in\mathcal{M}$, let $E_m$ be the edge set of the graph with sparsity pattern $m$. Let $\widehat{\Theta}_m^{(1)}$ be the constrained maximum likelihood estimator, defined by  
\begin{align}
\widehat{\Theta}_m^{(1)} = \underset{\Theta\in\mathcal{C}_m}{\operatorname{argmax}} \;\{\log\det(\Theta)-\emph{\text{tr}}(\widehat\Sigma_{n_1}^{(1)}\Theta)\},
\end{align}
where
\begin{align}
\mathcal{C}_m&=\{\Theta\succ0:\theta_{ij}=0\emph{\text{ for any }}(i,j)\not\in E_m\emph{\text{ and }}i\not=j\},
\end{align}
and $\widehat\Sigma_{n_1}^{(1)}$ denotes the empirical covariance matrix using the first subsample $\mathcal{D}_{n_1}^{(1)}$.
\end{definition} 

Notice that each individual estimator maximizes the log-likelihood under the constraint that some pre-defined subset of the parameters are zero. If $n_1\geq q_m$ where $q_m$ is the maximal clique size of a minimal chordal cover of the graph with edge set ${E}_m$, estimator $\widehat{\Theta}_m^{(1)}$ exists and is unique \cite{Uhler12}.
The following relationships hold regarding $\widehat{\Theta}_m^{(1)}$ and its inverse $(\widehat{\Theta}_m^{(1)})^{-1}$:
\begin{align}
[\widehat{\Theta}_m^{(1)}]_{ij} &= 0,\;\forall(i,j)\not\in{E}_m,\text{ and} \\
[(\widehat{\Theta}_m^{(1)})^{-1}]_{ij} &= [\widehat\Sigma_{n_1}^{(1)}]_{ij},\;\forall(i,j)\in{E}_m\cup\{(i,i),i=1,\ldots,p\}.
\end{align}
Indeed we can drive the above properties via the Lagrange form, where we add Lagrange constants for all missing edges of ${E}_m$
\begin{align}
\label{cmle}
l_{n_1,m}(\Theta):=\log\det(\Theta)-\text{tr}(\widehat\Sigma_{n_1}^{(1)}\Theta)-\sum_{(i,j)\not\in{E}_m}\gamma_{ij}\theta_{ij},
\end{align}
where $\gamma_{ij}$ is a Lagrange constant. The gradient equation for maximizing (\ref{cmle}) can be written as
\begin{align}
\Theta^{-1}-\widehat\Sigma_{n_1}^{(1)}-\Gamma=0,
\end{align}
where $\Gamma$ is a matrix of Lagrange parameters with nonzero values for all pairs with edges absent. It is an equality-constrained convex optimization problem, and a number of methods have been proposed for solving it, for example, in Hastie et~al.~\cite{Hastie09}. 

Now we define the aggregation estimator, which linearly combines a set of individual estimators $\widehat{\Theta}_m^{(1)}$ for $m\in\mathcal{M}$.

\begin{definition}
\label{maindef}
Let $\widehat{\Theta}_{\emph{\text{gES}}}^{(1,2)}$ be the graphical Exponential Screening (gES) estimator that linearly combines a set of individual estimators $\widehat{\Theta}_m^{(1)}$, defined by
\begin{align}
\widehat{\Theta}_{\emph{\text{gES}}}^{(1,2)}=\sum_{m\in\mathcal{M}}v_m^{(1,2)}\widehat{\Theta}_m^{(1)},
\end{align}
where the superscripts denote which subsamples are used for constructions, and the weights
\begin{align}
&v_m^{(1,2)}: =
\frac{\exp\{(n_2/2)(\emph{\text{logdet}}(\widehat{\Theta}_m^{(1)})-\emph{\text{tr}}(\widehat{\Theta}_m^{(1)}\widehat{S}_{n_2}^{(2)}))\}\pi_m}{\sum_{m'\in\mathcal{M}}\exp\{(n_2/2)(\emph{\text{logdet}}(\widehat{\Theta}_{m'}^{(1)})-\emph{\text{tr}}(\widehat{\Theta}_{m'}^{(1)}\widehat{S}_{n_2}^{(2)}))\}\pi_{m'}}.
\end{align}
Here, $S_{n_2}^{(2)}$ is an estimator of $\Sigma$, and $\pi_m$ is a (prior) probability distribution on the set of sparsity pattern $\mathcal{M}$, defined by
\begin{align}
\pi_m:= \frac{1}{H}\left(\frac{|m|_1}{e p(p-1)}\right)^{|m|_1},
\end{align}
where $H$ is a normalization factor to ensure that $\pi_m$ add up to one.
\end{definition}

Observe that the gES estimator is a linear combination of the set of individual estimators $\widehat{\Theta}_m^{(1)}$ with weights $v_m^{(1,2)}$. It is indeed a convex combination since the weights add up to one.

A natural choice of $S_{n_2}^{(2)}$ would be the empirical covariance matrix $\widehat{\Sigma}_{n_2}^{(2)}$. In this scenario, ignoring the prior distribution $\pi_m$, the individual weight is proportional to the likelihood of estimator $\widehat{\Theta}_m^{(1)}$ evaluated on the second subsample. Thus, the higher the predictive ability, the more weighting will be put on the corresponding individual estimator. However, as is shown in the next section, this would lead to a deterioration of convergence rate in high-dimensional settings. Instead, we can use the hard thresholding estimator proposed in \citet{Bickel08}. To be more specific, the thresholding estimation of $\widehat{\Sigma}_{n_2}^{(2)}$ thresholded at $\gamma$ is defined by 
\begin{align}
S_{n_2}^{(2)}=T_\gamma(\widehat\Sigma_{n_2}^{(2)}):=\{\hat\sigma_{ij}^{(2)}\cdot\mathbf{1}(|\hat\sigma_{ij}^{(2)}|\geq \gamma)\}_{1\leq i,j\leq p},
\end{align}
where $\hat\sigma_{ij}^{(2)}$ is the $(i,j)$th element of $\widehat{\Sigma}_{n_2}^{(2)}$. 
In practice, we can apply the following procedure for the problem of threshold selection \cite{Bickel08}: we split the second subsample randomly into two pieces of size $n_2(1-1/\log n_2)$ and $n_2/\log n_2$, respectively, and repeat this $B$ times. Let $\widehat\Sigma_{1,v}^{(2)}$ and $\widehat\Sigma_{2,v}^{(2)}$ be the empirical covariance matrices based on the two pieces, from the $v$th split. Then the thresholding parameter $\gamma$ is determined by
\begin{align}
\gamma=\underset{\gamma'}{\text{argmin}}\left\{\frac{1}{B}\sum_{v=1}^B\|T_{\gamma'}(\widehat\Sigma_{1,v}^{(2)})-\widehat\Sigma_{2,v}^{(2)}\|_F^2\right\}.
\end{align}

In Definition~\ref{maindef}, we also incorporate a deterministic factor $\pi_m$ into the weighted averaging to account for (prior) model complexity, in a manner that facilitates desirable risk properties \cite{Leung06, Rigollet11}. Here, low-complexity models are favored. Alternatively, if the set of sparsity patterns $\mathcal{M}=\{0, 1\}^{p(p-1)/2}$, the following deterministic factor specifies a uniform distribution on the cardinality of the subset
and a conditionally uniform distribution on the subsets of that size:
\begin{align}
\pi_m^{\text{U}}=\left[\left(p(p-1)/2+1\right){p(p-1)/2 \choose |m|_1}\right]^{-1}.
\end{align}
In addition, a simple way is to choose a flat prior, where we set $\pi_m^\text{F}=\pi_{m'}^\text{F}$ for any $m,m'\in\mathcal{M}$.

We show in the next section that, under some technical conditions, the risk of the gES estimator is bounded by the risk of the best individual estimator plus a low price for aggregating.

\subsection{Approximation algorithm}

To implement the estimating procedure, note that exact computation of the aggregation estimator might require the calculation of as many as $2^{p(p-1)/2}$ individual estimators. In many cases this number could be extremely large, and we must make a numerical approximation. Observing that the aggregation estimator is actually the expectation of a random variable that has a probability mass proportional to $v_m^{(1,2)}$ on individual estimator $\widehat{\Theta}_m^{(1)}$ for $m\in\mathcal{M}$, then the Metropolis-Hastings algorithm can be exploited to provide such an approximation. The detail algorithm is shown in Algorithm~\ref{alg}.

\begin{algorithm}
\label{alg}
\vskip 0.1in
If the set of sparsity patterns $\mathcal{M}=\{0, 1\}^{p(p-1)/2}$, then the gES estimator can be approximated by running a Metropolis-Hastings algorithm on a $p(p-1)/2$-dimensional hypercube:
\vspace{-\topsep}
\begin{itemize}
\item[(S1)] Initialize $m_t=\{0\}^{\frac{p(p-1)}{2}}, t=0$; 
\item[(S2)] For each $t\geq 0$, generate $m_t'$ with the uniform distribution on the neighbours of $m_t$;
\item[(S3)] Generate an $[0,1]$-uniformly distributed number $r$;
\item[(S4)] Put $m_{t+1} \leftarrow m_t'$, if $r < \min\{1,v_{m_t'}^{(1,2)}/v_{m_t}^{(1,2)}\}$; otherwise, $m_{t+1} \leftarrow m_t;$
\item[(S5)] Compute $\widehat\Theta_{m_{t+1}}^{(1)}$. Stop if $t>T_0+T$; 
otherwise, update $t \leftarrow t+1$ and go to step (S2).
\end{itemize}
\vspace{-\topsep}
Then we can approximate $\widehat\Theta_{\emph{\text{gES}}}^{(1,2)}$ by 
\begin{align}
\widehat{\widehat{\Theta}_{\emph{\text{gES}}}^{(1,2)}}=\frac{1}{T}\sum_{t=T_0+1}^{T_0+T}\widehat\Theta_{m_t}^{(1)},
\end{align}
where $T_0$ and $T$ are positive integers.
\vskip 0.1in
\end{algorithm}
Here, the neighbours of $m_t$ consists of all the sparsity patterns with a Manhattan distance of one to $m_t$. The following proposition shows that the resulting Markov chain ensures the ergodicity. 
\begin{proposition}
The Markov chain $\{m_t\}_{0\leq t\leq T_0+T}$ generated by Algorithm~\ref{alg} satisfies
\begin{align}
\lim_{T\to \infty}\frac{1}{T}\sum_{t=T_0+1}^{T_0+T}\widehat\Theta_{m_t}^{(1)}=\sum_{m\in\mathcal{M}}v_m^{(1,2)}\widehat{\Theta}_m^{(1)},\;\text{almost surely}.
\end{align}
\end{proposition}

The proof is straightforward as the Markov chain is clearly $v^{(1,2)}$-irreducible.

The Metropolis-Hastings algorithm incorporates a trade-off between prediction and sparsity to decide whether to add or discard an edge. Observe that the gES estimator would always estimate a precision matrix in which all the elements are nonzero, since all possible individual estimators are linearly mixed. However, the Metropolis-Hastings algorithm would lead to a sparse precision matrix estimation as in the regression case \cite{Rigollet11}.

\section{Theoretical Properties}
\label{theory}

In this section, we show that under some technical conditions, the risk of the gES estimator is bounded by the risk of the best individual estimator in the dictionary plus a low aggregating price. 

Following the notations in \citet{Bickel08}, we define the uniformity class of covariance matrices invariant under permutations by
\begin{align}
\mathcal{U}(q,\delta,M)=\Bigl\{\Sigma\in\mathbb{R}^{p\times p}:\;\Sigma\succ0,\;\sigma_{ii}\leq M,\;
\sum_{j=1}^p|\sigma_{ij}|^q\leq \delta,\;\text{for all }i\Bigr\},
\end{align}
for $0\leq q<1$, where $M$ and $\delta$ are constants.

For any estimator $\widehat\Theta\succ0$, we define the risk function
\begin{align}
R(\widehat\Theta)=\text{tr}(\widehat\Theta\Sigma)-\text{logdet}(\widehat\Theta).
\end{align}
Note that $\Sigma$ is the true covariance matrix. Consider the Kullback-Leibler (KL) divergence
\begin{align}
\text{KL}(\widehat\Theta)=R(\widehat\Theta)-R(\Theta)\geq0.
\end{align}

For each individual estimator corresponding to $m\in\mathcal{M}$, we assume $\widehat{\Theta}_m^{(1)}$ exists and is unique. The following proposition relates the KL risk of the aggregation estimator $\widehat{\Theta}_{{\text{gES}}}^{(1,2)}$ to the KL risks of individual estimators $\widehat{\Theta}_{m}^{(1)}$.

\begin{proposition} 
\label{prop1}
The gES estimator $\widehat{\Theta}_{\emph{\text{gES}}}^{(1,2)}$ in Definition~\ref{maindef} satisfies the following inequality
\begin{align}
\emph{\text{KL}}(\widehat{\Theta}_{\emph{\text{gES}}}^{(1,2)})\leq &
\min_{m\in\mathcal{M}}\Bigl \{\emph{\text{KL}}(\widehat{\Theta}_{m}^{(1)})+\frac{2}{n_2}\log\frac{1}{\pi_{m}}+
\emph{\text{tr}}((\widehat{\Theta}_{m}^{(1)}-\widehat{\Theta}_{\emph{\text{gES}}}^{(1,2)})(\widehat{S}_{n_2}^{(2)}-\Sigma))\Bigr\} \\
\leq&\min_{m\in\mathcal{M}}\Bigl \{\emph{\text{KL}}(\widehat{\Theta}_{m}^{(1)})+\frac{2}{n_2}\log\frac{1}{\pi_{m}}+
\emph{\text{tr}}(\widehat{\Theta}_{m}^{(1)}(\widehat{S}_{n_2}^{(2)}-\Sigma))
+\|\widehat{S}_{n_2}^{(2)}-\Sigma\|\cdot\emph{\text{tr}}(\widehat{\Theta}_{\emph{\text{gES}}}^{(1,2)}).
\end{align}
\end{proposition}

It is shown in \citet{Vershynin12} that under some conditions
\begin{align}
\|\widehat{\Sigma}_{n_2}^{(2)}-\Sigma\|=O_P\left(\sqrt{\frac{p}{n_2}}\right),
\end{align}
thus if we use the empirical covariance matrix as $\widehat{S}_{n_2}^{(2)}$, Proposition~\ref{prop1} implies that this would lead to a deterioration of convergence rate in high dimensions. Then we choose to use the thresholded covariance matrix estimation in \citet{Bickel08}.

We consider the following assumption for further analysis of the remainder term.
\begin{assumption}
\label{asm}
The set of sparsity patterns $\mathcal{M}$ satisfies the following condition
\begin{align}
\max_{m\in\mathcal{M}}{\emph{\text{tr}}}(\widehat{\Theta}_m^{(1)})=O_P(p).
\end{align}
\end{assumption}

This assumption ensures a fast convergence rate of the aggregation estimator. It is shown in Dahl et~al.~\cite{Dahl05} that the inverse of $\widehat{\Theta}_m^{(1)}$ is a solution to the following problem that maximizes the determinant of a symmetric positive definite matrix $Z$:
\begin{equation}
\begin{aligned}
& \underset{Z\succ 0}{\text{max}}
& & \text{logdet}\;Z, \\
& \text{  s.t.}
& & Z_{ij} = [\widehat{\Sigma}_{n_1}^{(1)}]_{ij},\;\forall(i,j)\in{E}_m\cup\{(i,i),i=1,\ldots,p\}.
\end{aligned}
\end{equation}
In general, note that for any two graphs $G$ and $G'$ with sparsity patterns $m$ and $m'$, respectively, if $G$ is a subgraph of $G'$, then the trace of $\widehat{\Theta}_{m'}^{(1)}$ would always be larger than that of $\widehat{\Theta}_{m}^{(1)}$. Thus the most dense graph in the dictionary would always be able to achieve the maximum trace among all individual estimators. This assumption claims that the diagonal entries of the true precision matrix $\Theta$ are well estimated by all individual estimators in the dictionary $\mathcal{M}$.

Let ${m}^*\in\mathcal{M}$ be the sparsity pattern that attains the minimum KL risk in the dictionary $\mathcal{M}$:
\begin{align}
{m}^*=\underset{m\in\mathcal{M}}{\text{argmin }}{\text{KL}}(\widehat{\Theta}_{m}^{(1)}).
\end{align}

The following theorem shows the oracle inequality that the aggregation estimator satisfies.

\begin{theorem}
\label{main}
Suppose Assumption~\ref{asm} hold. Uniformly on $\mathcal{U}(q,\delta, M)$, for sufficiently large $K$, if the thresholding parameter $\lambda=K\sqrt{\frac{\log p}{n_2}}$, and $\frac{\log p}{n_2}=o(1)$, then the gES estimator $\widehat{\Theta}_{\emph{\text{gES}}}^{(1,2)}$ satisfies
\begin{align}
\emph{\text{KL}}(\widehat{\Theta}_{\emph{\text{gES}}}^{(1,2)})-\min_{m\in\mathcal{M}}\emph{\text{KL}}(\widehat{\Theta}_{m}^{(1)})
=O_P\left(\frac{{s}^*\log p}{n_2}+p\left(\frac{\log p}{n_2}\right)^{(1-q)/2}\right),
\end{align}
where ${s}^*=|{m}^*|_1$ is the number of nonzero off-diagonal elements of $\widehat{\Theta}_{{m}^*}$.
\end{theorem}

This theorem yields a rate of convergence of the excess risk. In particular, if the set of sparsity patterns $\mathcal{M}$ also includes the true sparsity pattern $m^0$. Let $s^0$ be the number of nonzero off-diagonal elements of $\Theta$. It is shown in Zhou et~al.~\cite{Zhou11} that under some technical conditions
\begin{align}
{\text{KL}}(\widehat{\Theta}_{m^0}^{(1)})=O_P\left(\frac{(s^0+p)\log p}{n_2}\right).
\end{align}
Combine it with Theorem~\ref{main} and assume that $s^0=O(p)$, we obtain
\begin{align}
{\text{KL}}(\widehat{\Theta}_{{\text{gES}}}^{(1,2)})=O_P\left(p\left(\frac{\log p}{n_2}\right)^{(1-q)/2}\right),
\end{align}
for $0\leq q<1$.

\section{Experimental Results}
\label{num}

In this section, we provide empirical evidence to illustrate the usefulness of the proposed gES estimator and compare it with other state-of-the-art methods in parameter estimation and graph recovery using simulated and real datasets.

\subsection{Numerical simulations}
\begin{figure}[!h]
\begin{center}
\begin{tabular}{ccc}
\includegraphics[width=.25\textwidth]{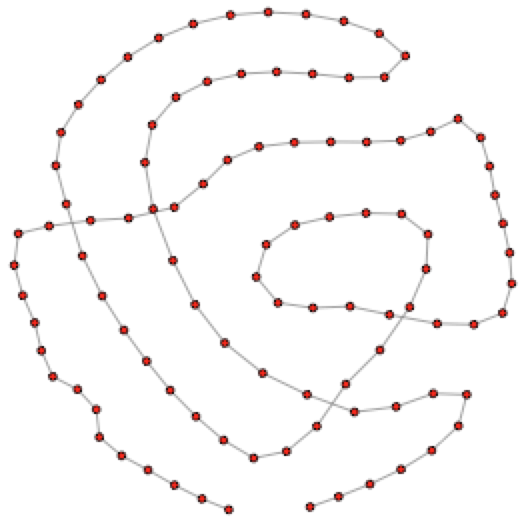} & $\quad$
\includegraphics[width=.25\textwidth]{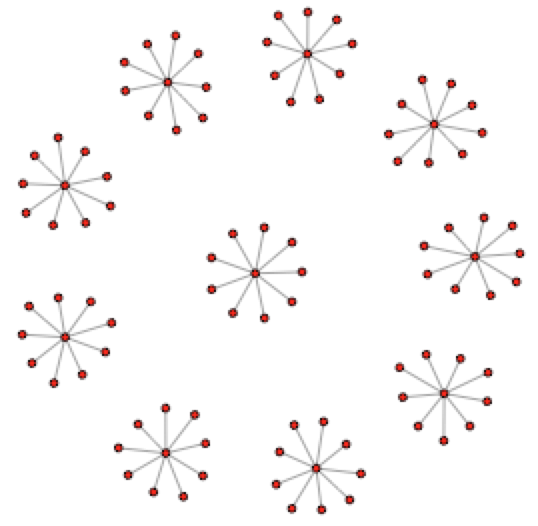} & $\quad$
\includegraphics[width=.25\textwidth]{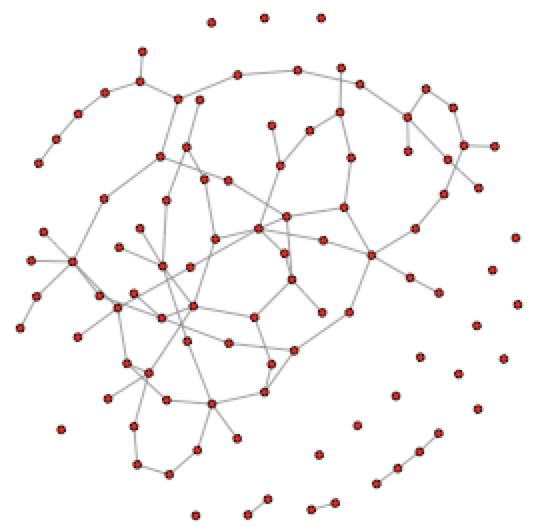} \\ \\
(a) AR & $\quad$ (b) Hub & $\quad$ (c) Random
\end{tabular}
\end{center}
\caption{{An illustration of the three graph patterns with $p=100$ nodes.}}
\label{fig:simulation}
\end{figure}

We generate synthetic datasets with sample size $n=200$ or $400$, and number of nodes $p=50,100$ or $200$. We use the following three models for simulating graphs and precision matrices. Figure~\ref{fig:simulation} displays a typical run of the generated graphs of the precision matrices when $p=100$.

\begin{itemize}
\item[(a)] ``AR'': The off-diagonal $(i,j)$th element of the adjacency matrix is set to be 1 if $|i-j|= 1$ and 0 otherwise;
\item[(b)] ``Hub'': The vertices are evenly partitioned into $p/10$ disjoint groups. Each group is associated with a center vertex $i$ in that group and the off-diagonal $(i,j)$th element of the adjacency matrix is set to 1 if $j$ also belongs to the same group as $i$ and 0 otherwise;
\item[(c)] ``Random": The off-diagonal $(i,j)$th element of the adjacency matrix is randomly set to be 1 with certain probability  and 0 otherwise.
\end{itemize}

We compare the proposed gES estimator with the graphical lasso (glasso) \cite{Friedman08, Yuan07} and constrained $\ell_1$ minimization estimator (CLIME) \citep{Cai11} with the tuning parameters determined by 10-fold cross-validation. We also consider the method of multiple testing of hypotheses about vanishing partial correlation coefficients \cite{Drton04, Drton07}, where the family-wise error rates for incorrect edge inclusion are controlled by Bonferroni correction at level $\alpha=0.05$. We refer this method as ``pcorTest'' in this paper. For the gES estimator, the full dataset is random partitioned into two subsamples with equal size, while for other methods, the graphs are estimated using the full dataset in order to make the comparison fair.

Notice that when $p$ is large, the Metropolis-Hastings algorithm for the gES estimator might take a large number of iterations to generate a good random tour over the hypercube with as many as $p(p-1)/2$ dimensions, and thus could become computationally unsuitable. In this case, we need to approximate the model space and select a candidate set of edges before the execution of the Metropolis-Hastings algorithm. We identify a candidate set of edges by applying some pre-screening method to the first subsample $\mathcal{D}_{n_1}^{(1)}$. Let the ordered list $\mathcal{Q}_p\subset\mathcal{S}_p$ be the set of selected edges. Then our aggregation process is constructed on this subset of edges. Then the set of sparsity patterns is given by
\begin{align}
\mathcal{M}=\prod_{k=1}^{p(p-1)/2}\mathcal{C}_k,
\end{align}
where
\begin{align}
\mathcal{C}_k: = \begin{cases} \{0,1\}\quad &\text{if }(i_k,j_k)\in\mathcal{Q}_p\text{, where }(i_k,j_k)\text{ is the }k\text{th}
\text{ element of }\mathcal{S}_p\\ \{0\}\quad &\text{otherwise} \end{cases}.
\end{align}
In practice where $p$ is large, the Metropolis-Hastings algorithm introduced in Algorithm~\ref{alg} will be applied to this reduced set of sparsity patterns instead of the set of all possible edges $\{0, 1\}^{p(p-1)/2}$. Many pre-screening methods could be considered here, for instance, the glasso. Note that we prefer to choose a small regularization parameter to prevent any true edge being ruled out in this stage. An alternative algorithm could be based on the $\ell_l$ regularization paths from the glasso method. Let $\mathcal{B}$ be a set of regularization parameters. For each $\beta\in\mathcal{B}$, let $m_\beta$ be the sparsity pattern resulted from glasso with tuning parameter $\beta$. Then the aggregation method is constructed on the set of sparsity patterns $\mathcal{M}=\{m_\beta:\beta\in\mathcal{B}\}$. However, in practical applications, the Metropolis-Hastings algorithm based on the $\ell_l$ regularization paths always concentrations on a single model after convergence, and generally does not perform well.

For any estimator $\widehat\Theta\succ0$, we use the following criteria for the comparisons.
\vspace{-\topsep}
\begin{itemize}
\item[(1)] Squared Frobenius norm error:
\begin{align}
\text{Frobenius}=\|\widehat\Theta-\Theta\|_F^2;
\end{align}
\item[(2)] Kullback-Leibler loss: 
\begin{align}
\text{KL}&=-\log\text{det}(\widehat{\Theta})+\text{trace}(\widehat{\Theta}\Sigma)-(-\log\text{det}(\Theta)+p);
\end{align}
\item[(3)] Precision: the number of correctly estimated edges divided by the total number of edges in the estimated graph. Let $G=(V, E)$ be a $p$-dimensional graph and let $\widehat{G}=(\widehat{V},\widehat{E})$ be an estimated graph. We define
\begin{align}
\text{Precision}=\frac{|\widehat{E}\cap E|}{|\widehat{E}|};
\end{align}
\item[(4)] Recall: the number of correctly estimated edges divided by the total number of edges in the true graph, defined by
\begin{align}
\text{Recall}=\frac{|\widehat{E}\cap E|}{|E|};
\end{align}
\item[(5)] $F_1$-score: a weighted average of the precision and recall, defined by 
\begin{align}
F_1\text{-score}=2\cdot\frac{\text{Precision}\cdot\text{Recall}}{\text{Precision}+\text{Recall}},
\end{align}
where an $F_1$-score reaches its best value at 1 and worst score at 0. 
\end{itemize}

\begin{table*}[!ht]
\caption{{Quantitative comparison of different methods on simulated data; Averaged quantities with their standard errors (in parentheses) are reported.}}
\vspace{0.2cm}
\label{tab:simresult}
\centering
\resizebox{9.22cm}{!}{
\begin{tabular}{llrrccc}\toprule
& & \multicolumn{5}{c}{$(n,p)=(200,50)$} \\
\cmidrule(l){3-7}
Model & Estimator & Frobenius\; & KL$\qquad$ & Precision & Recall & $F_1$-score \\
\midrule
AR& gES & 6.06 (0.97) & 1.61 (0.29) & 0.73 (0.05) & 0.97 (0.02) & 0.83 (0.03) \\
& pcorTest & 5.91 (1.48) & 1.73 (0.49) & 0.99 (0.01) & 0.89 (0.04) & 0.94 (0.02) \\
& glasso & 6.39 (0.82) & 1.51 (0.14) & 0.21 (0.03) & 1.00 (0.00) & 0.35 (0.04) \\
& CLIME & 5.46 (0.76) & 1.77 (0.25) & 0.15 (0.02) & 1.00 (0.00) & 0.26 (0.03) \\
\midrule
Hub& gES & 6.29 (1.57) & 1.44 (0.31) & 0.65 (0.06) & 0.97 (0.02) & 0.78 (0.04) \\
& pcorTest & 26.56 (3.77) & 6.02 (0.67) & 0.98 (0.02) & 0.53 (0.06) & 0.69 (0.05) \\
& glasso & 16.43 (1.84) & 1.53 (0.13) & 0.18 (0.02) & 1.00 (0.00) & 0.31 (0.03) \\
& CLIME & 7.34 (0.99) & 3.03 (0.43) & 0.14 (0.01) & 1.00 (0.00) & 0.24 (0.02) \\
\midrule
Random& gES & 5.94 (1.72) & 1.72 (0.46) & 0.73 (0.05) & 0.91 (0.06) & 0.81 (0.04) \\
& pcorTest & 8.55 (3.63) & 2.45 (0.93) & 0.99 (0.02) & 0.72 (0.13) & 0.83 (0.09) \\
& glasso & 7.09 (1.18) & 1.31 (0.16) & 0.20 (0.03) & 1.00 (0.01) & 0.33 (0.04) \\
& CLIME & 4.59 (0.91) & 1.68 (0.32) & 0.13 (0.02) & 1.00 (0.00) & 0.23 (0.03) \\
\midrule
\midrule
& & \multicolumn{5}{c}{$(n,p)=(200,100)$} \\
\cmidrule(l){3-7}
Model & Estimator & Frobenius\; & KL$\qquad$ & Precision & Recall & $F_1$-score \\
\midrule
AR& gES & 13.94 (2.58) & 3.10 (0.55) & 0.84 (0.03) & 0.98 (0.02) & 0.90 (0.02)\\
& pcorTest & 17.48 (2.70) & 4.55 (0.72) & 0.91 (0.03) & 0.88 (0.03) & 0.89 (0.02) \\
& glasso & 20.89 (1.77) & 3.71 (0.21) & 0.17 (0.02) & 1.00 (0.00) & 0.29 (0.02) \\
& CLIME & 13.90 (1.47) & 4.79 (0.50) & 0.13 (0.01) & 1.00 (0.00) & 0.23 (0.02) \\
\midrule
Hub& gES & 14.27 (2.77) &  3.29 (0.58) & 0.77 (0.03) & 0.97 (0.02) & 0.86 (0.02) \\
& pcorTest & 72.07 (5.98) & 14.92 (0.93) & 0.81 (0.08) & 0.42 (0.04) & 0.55 (0.04) \\
& glasso & 38.56 (1.67) & 3.61 (0.17) & 0.15 (0.01) & 1.00 (0.00) & 0.27 (0.02) \\
& CLIME & 17.27 (2.14) & 8.48 (0.89) & 0.12 (0.01) & 1.00 (0.00) & 0.22 (0.02) \\
\midrule
Random& gES & 12.89 (3.05) & 3.94 (0.82) & 0.85 (0.04) & 0.88 (0.05) & 0.86 (0.04) \\
& pcorTest & 26.05 (7.44) & 7.15 (1.70) & 0.82 (0.06) & 0.58 (0.10) & 0.68 (0.08) \\
& glasso & 16.58 (2.24) & 3.04 (0.22) & 0.18 (0.02) & 1.00 (0.00) & 0.30 (0.04) \\
& CLIME & 10.39 (1.12) & 4.27 (0.50) & 0.11 (0.01) & 1.00 (0.00) & 0.20 (0.02) \\
\midrule
\midrule
& & \multicolumn{5}{c}{$(n,p)=(400,100)$} \\
\cmidrule(l){3-7}
Model & Estimator & Frobenius\; & KL$\qquad$ & Precision & Recall & $F_1$-score \\
\midrule
AR& gES & 5.54 (1.12) & 1.14 (0.18) & 0.82 (0.03) & 1.00 (0.00) & 0.90 (0.02) \\
& pcorTest & 2.46 (0.30) & 0.51 (0.06) & 0.99 (0.01) & 1.00 (0.00) & 1.00 (0.00) \\
& glasso & 12.17 (0.95) & 1.93 (0.10) & 0.17 (0.01) & 1.00 (0.00) & 0.29 (0.02) \\
& CLIME & 7.55 (0.58) & 2.08 (0.21) & 0.14 (0.01) & 1.00 (0.00) & 0.25 (0.02) \\
\midrule
Hub& gES & 5.27 (1.12) & 1.16 (0.18) & 0.82 (0.03) & 0.99 (0.01) & 0.90 (0.02) \\
& pcorTest & 11.20 (2.66) & 2.96 (0.69) & 0.99 (0.01) & 0.90 (0.03) & 0.94 (0.02) \\
& glasso & 23.76 (1.76) & 1.91 (0.09) & 0.15 (0.02) & 1.00 (0.00) & 0.26 (0.03) \\
& CLIME & 8.31 (0.78) & 3.20 (0.27) & 0.13 (0.00) & 1.00 (0.00) & 0.23 (0.01) \\
\midrule
Random& gES & 5.46 (0.85) & 1.98 (0.33) & 0.84 (0.03) & 0.94 (0.04) & 0.88 (0.03) \\
& pcorTest & 6.53 (1.76) & 2.42 (0.65) & 0.99 (0.01) & 0.82 (0.06) & 0.90 (0.04) \\ 
& glasso & 6.48 (0.75) & 1.69 (0.15) & 0.19 (0.02) & 1.00 (0.00) & 0.32 (0.03) \\
& CLIME & 5.03 (0.54) & 1.95 (0.17) & 0.14 (0.02) & 1.00 (0.00) & 0.24 (0.02) \\
\midrule
\midrule
& & \multicolumn{5}{c}{$(n,p)=(400,200)$} \\
\cmidrule(l){3-7}
Model & Estimator & Frobenius\; & KL$\qquad$ & Precision & Recall & $F_1$-score \\
\midrule
AR& gES & 14.27 (2.42) & 2.99 (0.53) & 0.89 (0.02) & 0.99 (0.01) & 0.94 (0.01) \\
& pcorTest & 6.02 (1.13) & 1.36 (0.28) & 0.92 (0.02) & 1.00 (0.00) & 0.96 (0.01) \\
& glasso & 29.93 (2.37) & 4.62 (0.19) & 0.14 (0.02) & 1.00 (0.00) & 0.25 (0.03) \\
& CLIME & 16.33 (1.53) & 5.66 (0.61) & 0.10 (0.03) & 1.00 (0.00) & 0.18 (0.05) \\
\midrule
Hub& gES & 13.49 (2.63) & 2.89 (0.54) & 0.83 (0.04) & 0.99 (0.01) & 0.90 (0.03) \\
& pcorTest & 44.47 (4.72) & 11.25 (1.11) & 0.89 (0.03) & 0.79 (0.02) & 0.84 (0.02) \\
& glasso & 54.98 (1.84) & 4.45 (0.16) & 0.13 (0.00) & 1.00 (0.00) & 0.23 (0.01) \\
& CLIME & 19.65 (1.68) & 8.17 (0.50) & 0.14 (0.00) & 1.00 (0.00) & 0.25 (0.01) \\
\midrule
Random& gES & 10.43 (1.24) & 3.88 (0.51) & 0.95 (0.02) & 0.90 (0.04) & 0.92 (0.02) \\
& pcorTest & 13.21 (1.81) & 4.95 (0.74) & 0.87 (0.04) & 0.77 (0.06) & 0.81 (0.04) \\
& glasso & 11.67 (1.18) & 3.24 (0.20) & 0.16 (0.01) & 1.00 (0.00) & 0.28 (0.02) \\
& CLIME & 9.77 (0.97) & 3.56 (0.27) & 0.12 (0.01) & 1.00 (0.00) & 0.21 (0.02) \\
\bottomrule
\end{tabular}}
\end{table*}

Table~\ref{tab:simresult} shows the simulation results of the quantitative comparison of different methods, where we repeat the experiments 50 times and report the averaged values with their standard errors. We can see that the gES and CLIME estimators perform better than glasso and pcorTest in term of the squared Frobenius norm errors, while gES and glasso are better than CLIME and pcorTest regarding the Kullback-Leibler loss. For the comparison of graph structure recovery, the gES and pcorTest estimators outperform other methods.

\begin{figure}[!h]
\begin{center}
\vspace{-0.2cm}
\begin{tabular}{cc}
\includegraphics[width=.7\textwidth]{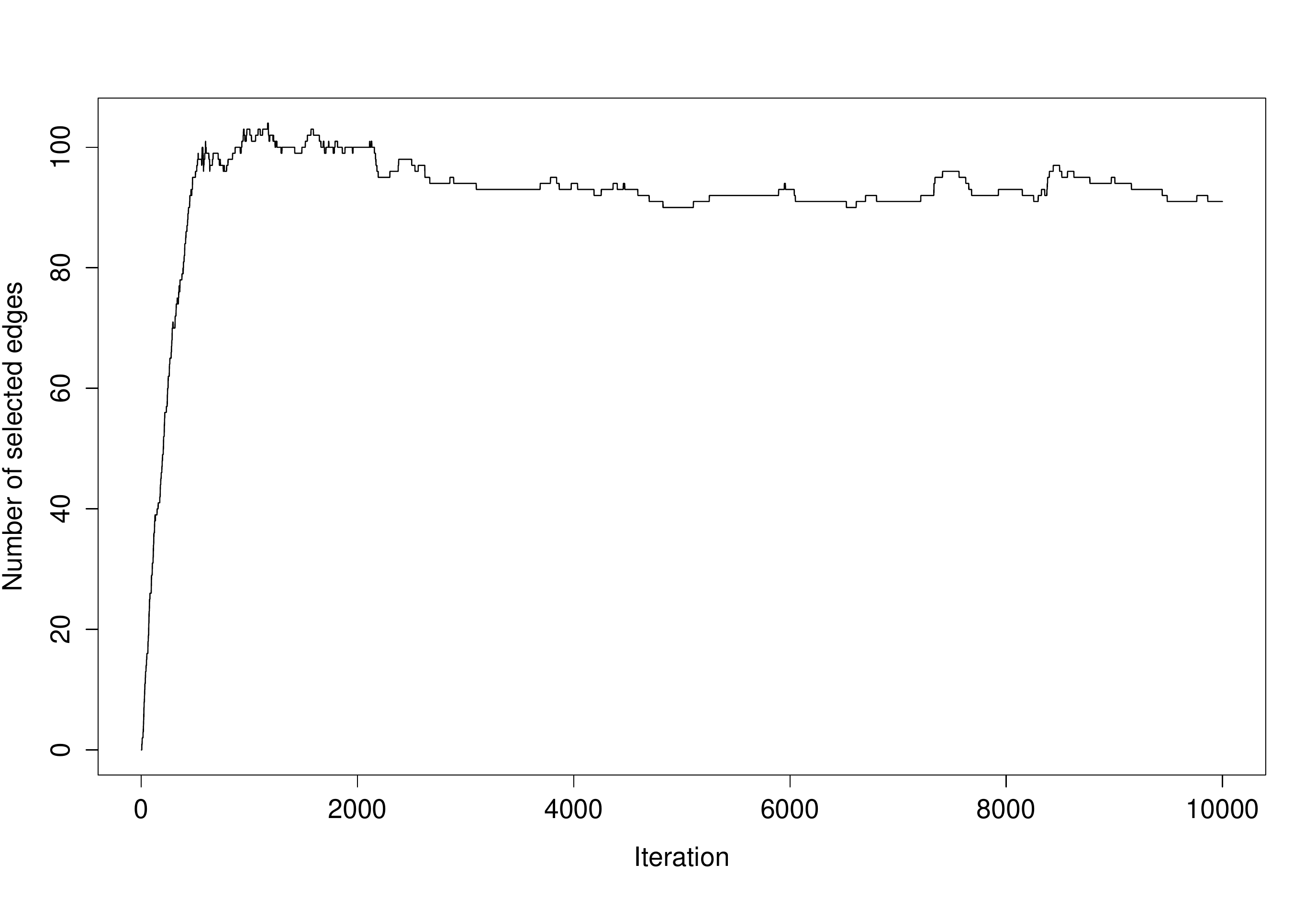} \\
\end{tabular}
\end{center}
\vspace{-0.4cm}
\caption{Number of selected edges by the Metropolis-Hastings algorithm as a function of iterations, given a typical run of the gES estimator for Hub graph on simulated data with $(n,p)=(400,100)$.}
\label{fig:mcmc}
\end{figure}

\begin{figure}[!h]
\vskip 0.2in
\begin{center}
\begin{tabular}{cc}
\includegraphics[width=.45\textwidth]{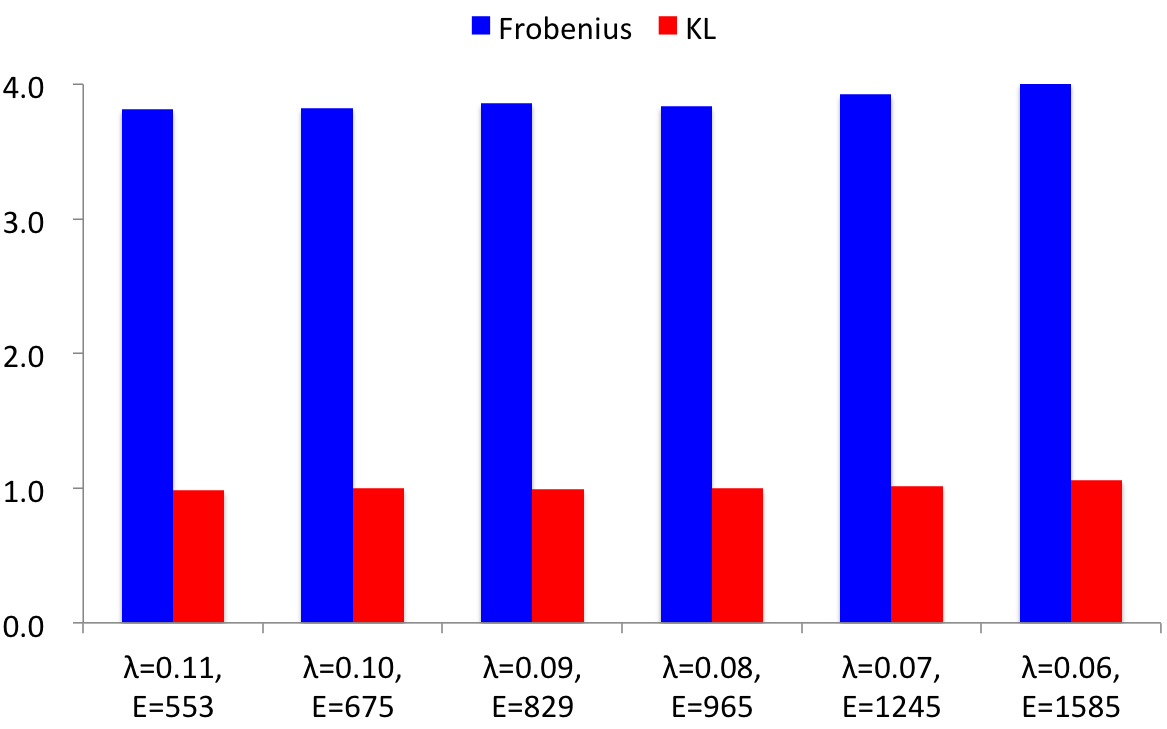} & $\quad$
\includegraphics[width=.45\textwidth]{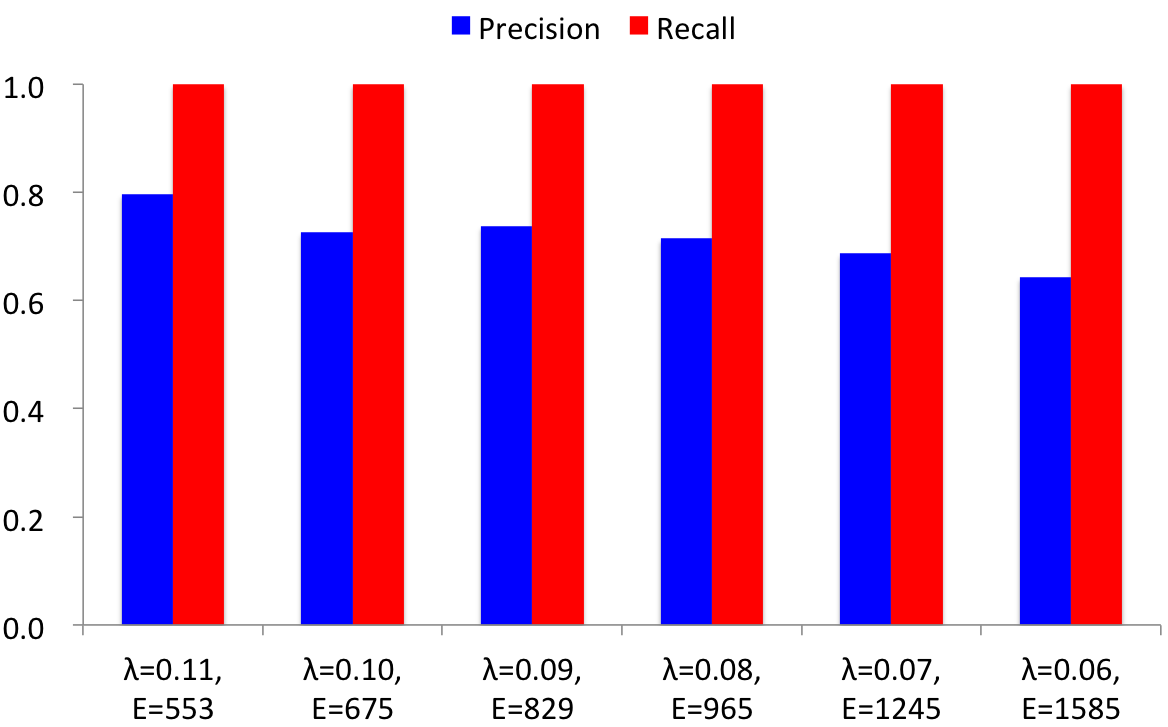}  \\
\end{tabular}
\end{center}
\caption{{Typical realization of the gES estimator for Hub graph on simulated data with $(n,p)=(400,100)$, varying the regularization parameter $\lambda$ in the pre-screening glasso; Number of selected candidate edges (marked as $E$) are also reported.}}
\label{fig:sensitivity}
\end{figure}

Figure~\ref{fig:mcmc} displays the evolution of the Metropolis-Hastings algorithm, showing evidence that the algorithm converges after 4000 iterations. Figure~\ref{fig:sensitivity} shows a typical realization of the gES method, varying the regularization parameter $\lambda$ in the pre-screening glasso, where we can see that the results for graphical aggregation are not quite sensitive to the pre-screening method.

The computational complexity of glasso \cite{Friedman08} which uses a row-by-row block coordinate algorithm is roughly $O(p^3)$. For the gES estimator, the complexity is roughly $O(L p^3)$, where $L=T+T_0$ is the number of iterations in the Metropolis-Hastings algorithm. Note that $L$ can be dramatically reduced if we keep track of and store the individual estimators to avoid duplicated computation of precision matrix with the same sparsity pattern.

\subsection{Analysis of microarray data}

In this study, we consider a real-world dataset based on Affymetrix GeneChip microarrays for the plant \emph{Arabidopsis thaliana} \cite{Wille04}. The sample size is $n = 118$. A nonparanormal transformation is estimated and the expression levels for each chip are replaced by their respective normal scores, subject to a Winsorized truncation \cite{Liu09}. A subset of $p=40$ genes from the isoprenoid pathway are chosen, and we study the associations among them using the proposed gES estimator and the glasso method with tuning parameter determined by cross-validation.

\begin{figure}[!h]
\begin{center}
\begin{tabular}{cc}
\includegraphics[width=.45\textwidth]{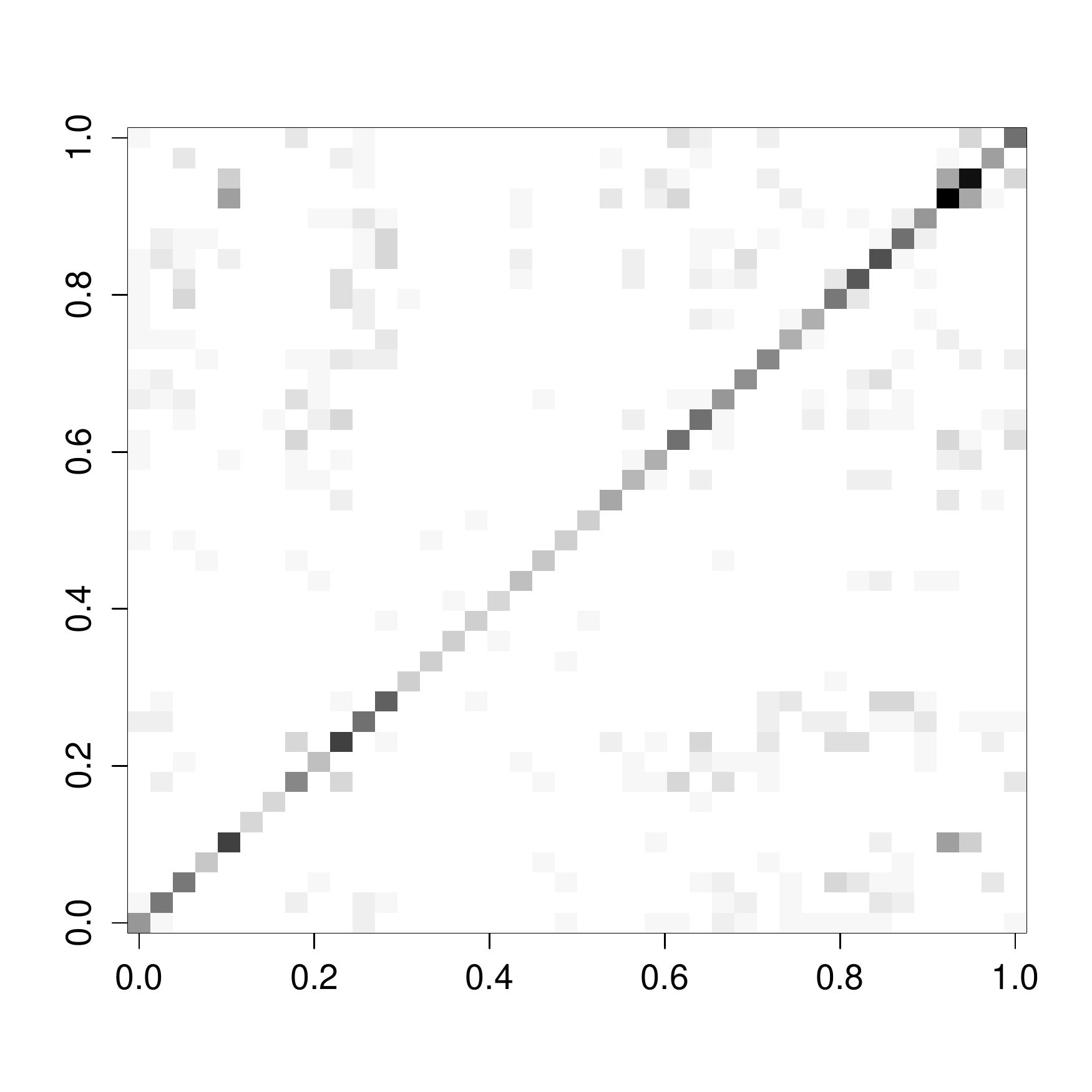} & $\quad$
\includegraphics[width=.45\textwidth]{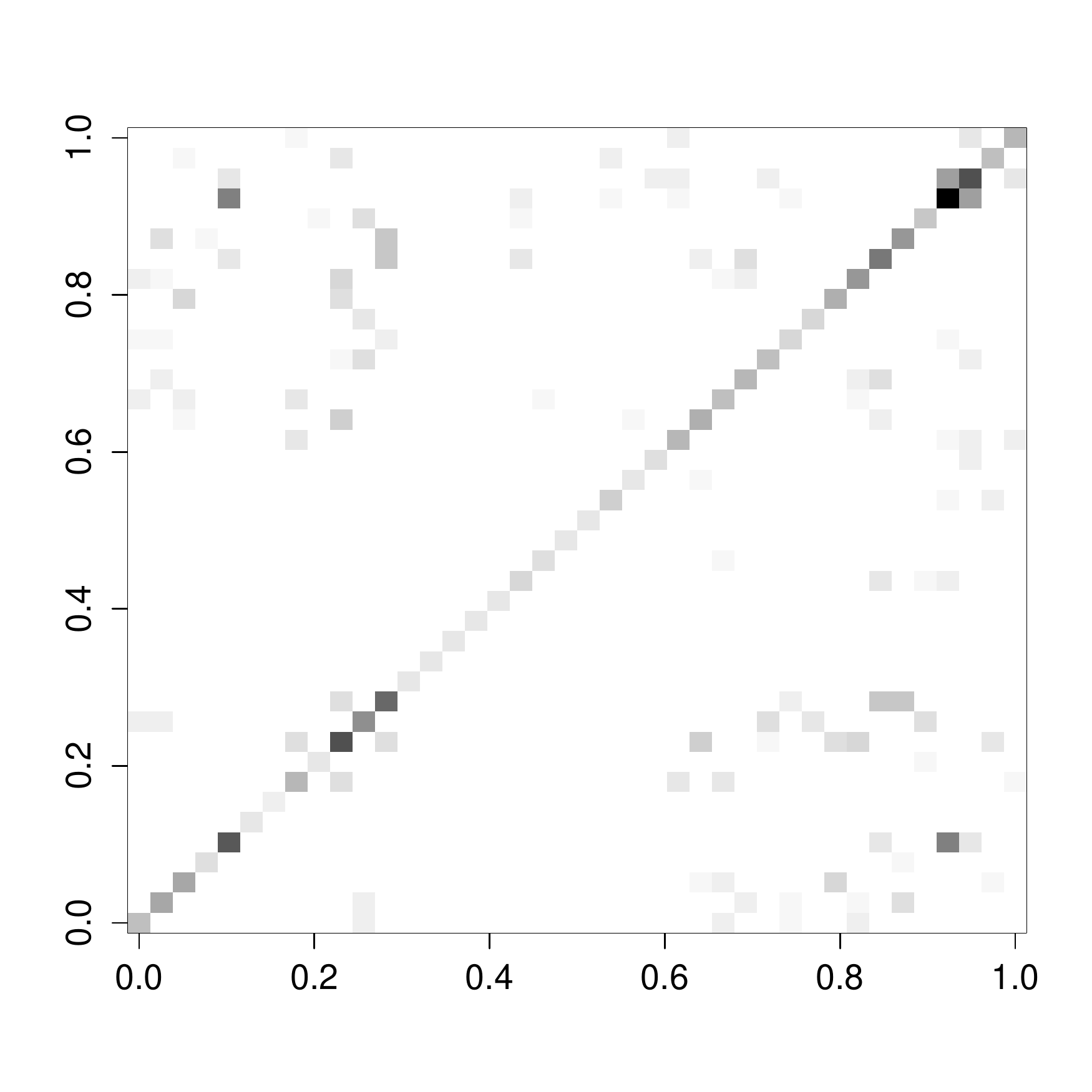} \\
(a) glasso & $\quad$ (b) gES \\
\end{tabular}
\end{center}
\caption{Grids of rectangles with gray scale corresponding to the absolute values in the estimated precision matrix for each method.}
\label{fig:image}
\end{figure}

The results show that glasso selects 378 edges and gES selects 111 edges.  Among those selected edges, 102 edges are identified by both methods. Figure~\ref{fig:image} shows grids of rectangles with gray scale corresponding to the absolute values in the estimated precision matrix for each method.

We also analyze a dataset on microarrays for the gene expression levels \cite{Nayak09}. Of the 4238 genes in immortalized B cells for $n=295$ normal individuals, we select $p=318$ genes that are associated with the phenotypes in genome-wide association studies. We study the estimated graphs obtained by the glasso and gES estimators. The expression levels for each gene are pre-processed by log-transformation and standardization.

\begin{figure}[!h]
\begin{center}
\begin{tabular}{cc}
\includegraphics[width=.45\textwidth]{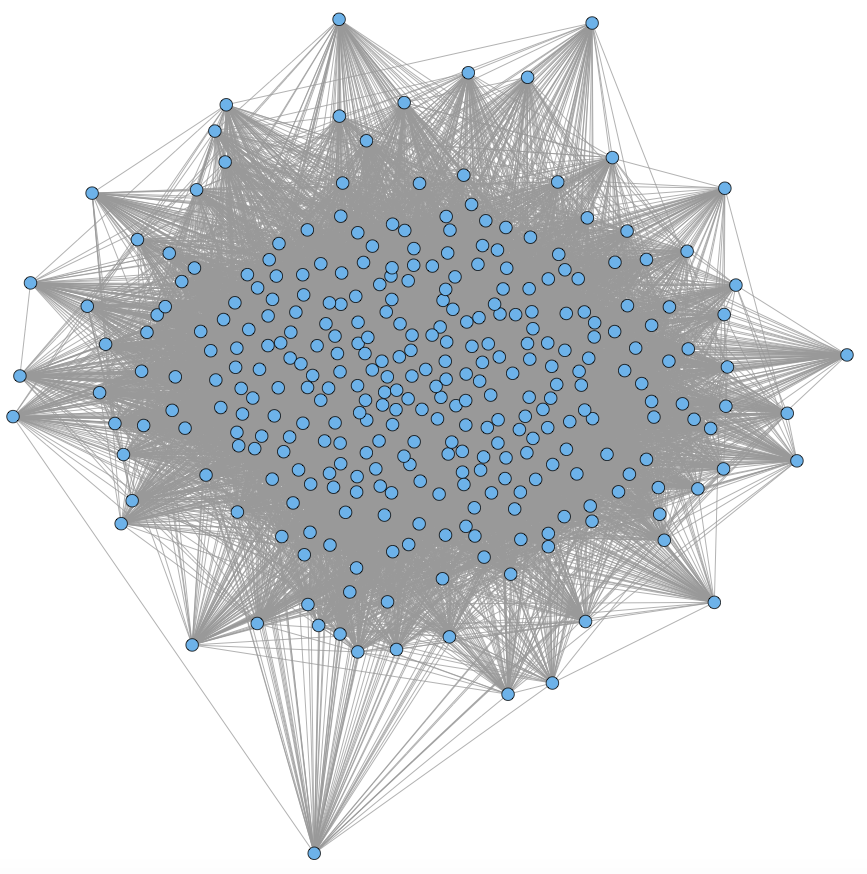} & $\quad$
\includegraphics[width=.45\textwidth]{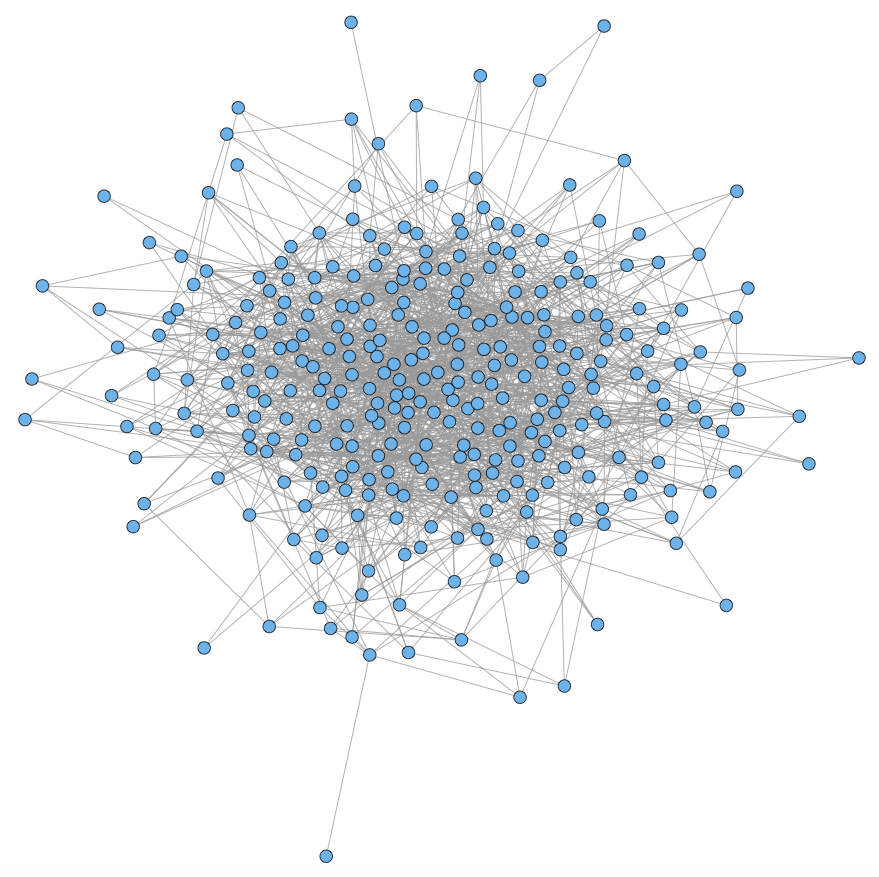} \\ \\
(a) glasso & $\quad$ (b) gES \\
\end{tabular}
\end{center}
\caption{Estimated graphs for microarray data example for immortalized human B cells.}
\label{fig:realgraph2}
\end{figure}

The results indicate that glasso selects 12514 edges and the gES estimator selects 1631 edges. Among those, 1542 edges are identified by both methods. Figure~\ref{fig:realgraph2} provides the estimated graphs.

\section{Conclusion}
\label{conclude}

In this paper, we propose a new aggregation method for estimating the precision matrix in Gaussian graphical models, by considering a convex combination of a suitable set of individual estimators with different underlying sparsity patterns. We investigate the risk of this aggregation estimator and show by an oracle that it is comparable to the risk of the best estimator based on a single graph. Experimental results validate the usefulness of our method in practice.

\section*{Acknowledgements}
This paper is based on part of my Ph.D dissertation. I want to thank my advisor, Dr. John Lafferty, for his help and guidance on this project.

\appendix

\section{Appendix}\label{app}
\subsection{Proof of Proposition~\ref{prop1}.}
\begin{proof}
For any $m\in\mathcal{M}$ and individual estimator $\widehat{\Theta}_m\succ0$, we obtain
\begin{align}
\KL(\widehat{\Theta}_m^{(1)})=\tr(\widehat{\Theta}_m^{(1)}\Sigma)-\text{logdet}(\widehat{\Theta}_m^{(1)})-p+\text{logdet}(\Theta).
\end{align}
Similarly, for the aggregation estimator $\widehat{\Theta}_{\text{gES}}^{(1,2)}\succ0$
\begin{align}
\KL(\widehat{\Theta}_{\text{gES}}^{(1,2)})=\tr(\widehat{\Theta}_{\text{gES}}^{(1,2)}\Sigma)-\text{logdet}(\widehat{\Theta}_{\text{gES}}^{(1,2)})-p+\text{logdet}(\Theta).
\end{align}
Based on the convexity of $\KL(\cdot)$, we obtain
\begin{align}
\label{ineq}
\KL(\widehat{\Theta}_{\text{gES}}^{(1,2)})\leq\sum_{m\in\mathcal{M}}v_m^{(1,2)}\KL(\widehat{\Theta}_{m}^{(1)}).
\end{align}
Let $\tilde{m}\in\mathcal{M}$ being any sparsity pattern attaining
\begin{align}
\label{tildem}
\min_{m\in\mathcal{M}}\left\{\KL(\widehat{\Theta}_{m}^{(1)})+\frac{2}{n_2}\log\frac{1}{\pi_{m}}+\tr(\widehat{\Theta}_{m}^{(1)}(\widehat{S}_{n_2}^{(2)}-\Sigma))\right\}.
\end{align}
Then according to the definition of $v_{m}^{(1,2)}$, we obtain
\begin{align}
\frac{v_{\tilde{m}}^{(1,2)}}{v_{m}^{(1,2)}}=&\frac{\exp\{(n_2/2)({\text{logdet}}(\widehat{\Theta}_{\tilde{m}}^{(1)})-{\text{tr}}(\widehat{\Theta}_{\tilde{m}}^{(1)}\widehat{S}_{n_2}^{(2)}))\}\pi_{\tilde{m}}}{\exp\{(n_2/2)({\text{logdet}}(\widehat{\Theta}_m^{(1)})-{\text{tr}}(\widehat{\Theta}_m^{(1)}\widehat{S}_{n_2}^{(2)}))\}\pi_m} \\
=&\frac{\pi_{\tilde{m}}}{\pi_m}\exp\{-(n_2/2)[\KL(\widehat{\Theta}_{\tilde{m}}^{(1)})-\KL(\widehat{\Theta}_{m}^{(1)})
+\tr((\widehat{\Theta}_{\tilde{m}}^{(1)}-\widehat{\Theta}_{m}^{(1)})(\widehat{S}_{n_2}^{(2)}-\Sigma))]\},
\end{align}
where the last equality follows from the fact that
\begin{align}
\KL(\widehat{\Theta}_{m}^{(1)})
=\tr(\widehat{\Theta}_m^{(1)}(\Sigma-\widehat{S}_{n_2}^{(2)}))-({\text{logdet}}(\widehat{\Theta}_m^{(1)})-{\text{tr}}(\widehat{\Theta}_m^{(1)}\widehat{S}_{n_2}^{(2)}))-p+\text{logdet}(\Theta).
\end{align}
Note that $\sum_{m\in\mathcal{M}}v_m^{(1,2)}=1$. Then the inequality (\ref{ineq}) can be written as
\begin{align}
\KL(\widehat{\Theta}_{\text{gES}}^{(1,2)})&\leq \KL(\widehat{\Theta}_{\tilde{m}}^{(1)})+\sum_{m\in\mathcal{M}}v_m^{(1,2)}(\KL(\widehat{\Theta}_{m}^{(1)})-\KL(\widehat{\Theta}_{\tilde{m}}^{(1)})) \\
\notag&\leq \KL(\widehat{\Theta}_{\tilde{m}}^{(1)})+\frac{2}{n_2}\sum_{m\in\mathcal{M}}v_m^{(1,2)}\log\frac{v_{\tilde{m}}^{(1,2)}}{v_{m}^{(1,2)}}+\frac{2}{n_2}\sum_{m\in\mathcal{M}}v_m^{(1,2)}\log\frac{\pi_m}{\pi_{\tilde{m}}} \\
&\quad+\sum_{m\in\mathcal{M}}v_m^{(1,2)}\tr(\widehat{\Theta}_{\tilde{m}}^{(1)}(\widehat{S}_{n_2}^{(2)}-\Sigma))
-\sum_{m\in\mathcal{M}}v_m^{(1,2)}\tr(\widehat{\Theta}_{m}^{(1)}(\widehat{S}_{n_2}^{(2)}-\Sigma)) \\
\notag&\leq \KL(\widehat{\Theta}_{\tilde{m}}^{(1)})+\frac{2}{n_2}\sum_{m\in\mathcal{M}}v_m^{(1,2)}\log\frac{\pi_m}{v_{m}^{(1,2)}}+\frac{2}{n_2}\sum_{m\in\mathcal{M}}v_m^{(1,2)}\log\frac{v_{\tilde{m}}^{(1,2)}}{\pi_{\tilde{m}}} \\
&\quad+\tr(\widehat{\Theta}_{\tilde{m}}^{(1)}(\widehat{S}_{n_2}^{(2)}-\Sigma))
-\sum_{m\in\mathcal{M}}v_m^{(1,2)}\tr(\widehat{\Theta}_{m}^{(1)}(\widehat{S}_{n_2}^{(2)}-\Sigma)).
\end{align}
According to the fact that
\begin{align}
\log v_{\tilde{m}}^{(1,2)}\leq 0,\text{ and }\sum_{m\in\mathcal{M}}v_m^{(1,2)}\log\frac{\pi_m}{v_{m}^{(1,2)}}\leq0,
\end{align}
we obtain
\begin{align}
\label{ineq2}
\KL(\widehat{\Theta}_{\text{gES}}^{(1,2)})\leq&\KL(\widehat{\Theta}_{\tilde{m}}^{(1)})+\frac{2}{n_2}\log\frac{1}{\pi_{\tilde{m}}}+\tr(\widehat{\Theta}_{\tilde{m}}^{(1)}(\widehat{S}_{n_2}^{(2)}-\Sigma))-\tr(\widehat{\Theta}_{\text{gES}}^{(1,2)}(\widehat{S}_{n_2}^{(2)}-\Sigma)).
\end{align}

It is shown in \citet{Wang86} that for any $p\times p$ real symmetric matrix $A$ and any $p\times p$ positive semidefinite matrix $B$
\begin{align}
\lambda_p(A)\tr(B)\leq\tr(AB)\leq\lambda_1(A)\tr(B),
\end{align}
where $\lambda_i(A)$ is the $i$th largest eigenvalue of $A$.

Following this property, we obtain
\begin{align}
\Bigl |\tr(\widehat{\Theta}_{\text{gES}}^{(1,2)}(\widehat{S}_{n_2}^{(2)}-\Sigma))\Bigr |
\leq\|\widehat{S}_{n_2}^{(2)}-\Sigma\|\cdot\tr(\widehat{\Theta}_{\text{gES}}^{(1,2)}).
\end{align}
Then we write the inequality (\ref{ineq2}) as
\begin{align}
\KL(\widehat{\Theta}_{\text{gES}}^{(1,2)})\leq&\KL(\widehat{\Theta}_{\tilde{m}}^{(1)})+\frac{2}{n_2}\log\frac{1}{\pi_{\tilde{m}}}+\tr(\widehat{\Theta}_{\tilde{m}}^{(1)}(\widehat{S}_{n_2}^{(2)}-\Sigma))+\|\widehat{S}_{n_2}^{(2)}-\Sigma\|\cdot\tr(\widehat{\Theta}_{\text{gES}}^{(1,2)}).
\end{align}
According to the definition of $\tilde{m}$ as in (\ref{tildem}), the proposition then follows.
\end{proof}

\vspace{5mm}
\subsection{Proof of Theorem~\ref{main}.}
\begin{proof}
Since ${m}^*\in\mathcal{M}$ is any sparsity pattern attaining $\min_{m\in\mathcal{M}}\KL(\widehat{\Theta}_{m}^{(1)}),$
then
\begin{align}
\text{KL}(\widehat{\Theta}_{\text{gES}}^{(1,2)})\leq&\text{KL}(\widehat{\Theta}_{{m}^*}^{(1)})+\frac{2}{n_2}\log\frac{1}{\pi_{{m}^*}}+\text{tr}(\widehat{\Theta}_{{m}^*}^{(1)}(\widehat{S}_{n_2}^{(2)}-\Sigma))+\|\widehat{S}_{n_2}^{(2)}-\Sigma\|\cdot\text{tr}(\widehat{\Theta}_{\text{gES}}^{(1,2)}) \\
=&\min_{m\in\mathcal{M}}{\text{KL}}(\widehat{\Theta}_{m}^{(1)})+\frac{2}{n_2}\log\frac{1}{\pi_{{m}^*}}+\text{tr}(\widehat{\Theta}_{{m}^*}^{(1)}(\widehat{S}_{n_2}^{(2)}-\Sigma))+\|\widehat{S}_{n_2}^{(2)}-\Sigma\|\cdot\text{tr}(\widehat{\Theta}_{\text{gES}}^{(1,2)}).
\end{align}
The normalization factor $H$ of the prior probability $\pi_m$ in Definition~\ref{maindef} satisfies the following inequality
\begin{align}
H&=\sum_{m\in\mathcal{M}}\left(\frac{|m|_1}{e p(p-1)}\right)^{|m|_1} \\
&\leq\sum_{k=0}^{p(p-1)/2}{p(p-1)/2 \choose k}\left(\frac{k}{e p(p-1)}\right)^k \\
&\leq\sum_{k=0}^{p(p-1)/2}\left(\frac{ep(p-1)/2}{k}\right)^k\left(\frac{k}{e p(p-1)}\right)^k \\
&\leq 2.
\end{align}
Then for any $m\in\mathcal{M}$ we have
\begin{align}
\log\left(\frac{1}{\pi_m}\right)&\leq|m|_1\log\left(\frac{e p(p-1)}{|m|_1\vee 1}\right)+\log 2.
\end{align}
Thus we obtain
\begin{align}
\label{p1}
\frac{2}{n_2}\log\frac{1}{\pi_{{m}^*}}=O_P\left(\frac{{s}^*\log p}{n_2}\right),
\end{align}
where ${s}^*=|{m}^*|_1$ is the number of nonzero off-diagonal elements of $\widehat{\Theta}_{{m}^*}$.

Note that
\begin{align}
\text{tr}(\widehat{\Theta}_{{m}^*}^{(1)}(\widehat{S}_{n_2}^{(2)}-\Sigma))
&\leq\|\widehat{S}_{n_2}^{(2)}-\Sigma\|\cdot\tr(\widehat{\Theta}_{{m}^*}^{(1)}) \\
&\leq\|\widehat{S}_{n_2}^{(2)}-\Sigma\|\cdot\max_{m\in\mathcal{M}}\tr(\widehat{\Theta}_m^{(1)}),
\end{align}
and
\begin{align}
\|\widehat{S}_{n_2}^{(2)}-\Sigma\|\cdot\tr(\widehat{\Theta}_{\text{gES}}^{(1,2)})
&=\|\widehat{S}_{n_2}^{(2)}-\Sigma\|\cdot\sum_{m\in\mathcal{M}}v_m^{(1,2)}\tr(\widehat{\Theta}_m^{(1)}) \\
&\leq\|\widehat{S}_{n_2}^{(2)}-\Sigma\|\cdot\max_{m\in\mathcal{M}}\tr(\widehat{\Theta}_m^{(1)}).
\end{align}
Given the assumptions of the theorem, it is shown in \citet{Bickel08} that
\begin{align}
\label{p3}
\|\widehat{S}_{n_2}^{(2)}-\Sigma\|=O_P\left(\left(\frac{\log p}{n_2}\right)^{(1-q)/2}\right).
\end{align}

Assumption~\ref{asm} provides a stochastic bound for $\max_{m\in\mathcal{M}}\tr(\widehat{\Theta}_m^{(1)})$. Combining it with the results of (\ref{p1}) and (\ref{p3}), the theorem follows.
\end{proof}

\bibliographystyle{plainnat}
\bibliography{gES}

\end{document}